\pdfoutput=1

\documentclass[11pt]{article}

\usepackage[final]{acl}

\usepackage{times}
\usepackage{latexsym}

\usepackage[T1]{fontenc}

\usepackage[utf8]{inputenc}

\usepackage{microtype}

\usepackage{inconsolata}

\usepackage{graphicx}
\usepackage{xcolor}
\usepackage{amsmath}
\usepackage{amssymb}
\usepackage{multirow}
\usepackage{rotating}
\usepackage{colortbl}

\title{How to Leverage Digit Embeddings to Represent Numbers?}

\author{Jasivan Alex Sivakumar \and Nafise Sadat Moosavi \\
  Department of Computer Science \\ University of Sheffield \\ United Kingdom\\
  \texttt{\{jasivakumar1|n.s.moosavi\}@sheffield.ac.uk}}

\begin{document}
\maketitle
\begin{abstract}
Within numerical reasoning, understanding numbers themselves is still a challenge for existing language models. Simple generalisations, such as solving 100+200 instead of 1+2, can substantially affect model performance \cite{sivakumar-moosavi-2023-fermat}. Among various techniques, character-level embeddings of numbers have emerged as a promising approach to improve number representation. However, this method has limitations as it leaves the task of aggregating digit representations to the model, which lacks direct supervision for this process. In this paper, we explore the use of mathematical priors to compute aggregated digit embeddings and explicitly incorporate these aggregates into transformer models. This can be achieved either by adding a special token to the input embeddings or by introducing an additional loss function to enhance correct predictions. We evaluate the effectiveness of incorporating this explicit aggregation, analysing its strengths and shortcomings, and discuss future directions to better benefit from this approach. Our methods, while simple, are compatible with any pretrained model, easy to implement, and have been made publicly available.\footnote{\url{https://github.com/jasivan/Number-Embeddings}}
\end{abstract}
\section{Introduction}
Numbers play an integral role in language \cite{thawani-etal-2021-representing}, and they are crucial across various domains such as finance \cite{chen-etal-2018-finance}, medicine \cite{jullien-etal-2023-semeval} or even sarcasm \cite{dubey-etal-2019-numbers}. Despite, large language models improving their capacity in many tasks, numerical reasoning still poses a challenge \citep{hong-etal-2024-stuck}. Recent advancements in enhancing numerical reasoning within language models have predominantly stemmed from using more extensive or higher-quality training datasets \cite{Li-etal-2022-semantic_augmentation, yu-etal-2023-metamath}, scaling up models \cite{lewkowycz-etal-2022-solving, kojima-etal-2022-large}, or integrating prompt-based strategies such as chain-of-thought reasoning \cite{wei-etal-2022-chain-of-thought, yue-etal-2024-mammoth}. The effectiveness of such methods is significantly amplified when applied in conjunction with larger model architectures. With smaller models, the improvement shown is often minimal, for example, \citet{wei-etal-2022-chain-of-thought} use of chain-of-thought on a 20B parameter model only showed a 2.5\% improvement on the MAWPS \cite{koncel-kedziorski-etal-2016-mawps} dataset whereas it jumps to 14.7\% with a 137B parameter model. In addition, many of these solutions are computationally expensive or inaccessible; we seek a low-cost approach that may have minimal impact on small-scale models but greater effects on larger models.

A key challenge for number understanding is that widely used tokenisation methods, like Byte-Pair Encoding (BPE) \cite{sennrich-etal-2016-neural}, work well for common words but not for numbers. Specifically, rarer numbers might be broken down into random and meaningless pieces. In light of this, digit tokenisation \cite{spithourakis-riedel-2018-numeracy} stands out for its simplicity and efficacy at representing numbers. This technique involves breaking down numbers into their individual digits, reducing vocabulary size and ensuring all decimal numbers can be accurately represented enhancing numerical reasoning abilities across various model architectures, tasks, and datasets \cite{geva-etal-2020-injecting, petrak-etal-2023-arithmetic, sivakumar-moosavi-2023-fermat}.  However, the aggregation of digit embeddings into a complete number representation is implicitly handled by the model, which raises the question: \textbf{Can explicit aggregation using mathematical priors improve numerical understanding?}

In this paper, we investigate this question by integrating a mathematically grounded aggregation of digit embeddings explicitly, rather than relying solely on the model's inherent capabilities. Our findings show that this aggregated digit embedding enhances performance on small-scale models by up to 6.17\% compared to our baseline without it, potentially leading to even greater improvements in larger models. However, the effectiveness of our integration strategy depends on the size, the architecture of the model used, and how these priors are integrated in the model. Our main contributions are as follows:
\begin{itemize}
    \item We propose a novel approach to number embedding that requires no changes to the model's architecture or additional pretraining, by showing that an aggregation is effective if it meets two criteria: (1) it distinguishes between distinct numbers, ensuring unique representations for each value, and (2) the aggregated embedding reflects natural numerical proximity.
    \item We explore two approaches for integrating our aggregation: adding a special token before the representation of individual digits to enhance input number representations, and incorporating an additional loss function to improve the representation of outputted digits.
\end{itemize}

\section{Related Work}
Numerical reasoning is the ability to interact with numbers using fundamental mathematical properties and model an area of human cognitive thinking \citep{saxton-etal-2018-analysing-NN-models}. Given a maths worded problem, for example ``Sarah has 5 apples and eats 2. How many apples does she have left?'', the model needs to interpret the relation between both the numbers and the text to then solve the problem by means of arithmetic operations \cite{ahn-etal-2024-survey}. Therefore, an accurate number representation is primordial to distinguish between different numbers and also predict an accurate answer. The literature focuses on five different areas to better represent numbers.

\subsection{Scaling}
Increasing the number of parameters of pretrained models has improved their numerical reasoning but it is still nowhere near perfect. For example, Minerva (540B) \cite{lewkowycz-etal-2022-solving} continues to struggle with greater than seven digit multiplication. Moreover, \citet{Frieder-etal-2023-Mathematical-Capabilities} found that very large models like ChatGPT and GPT4 are inconsistent when answering mathematical questions ranging from arithmetic problems to symbolic maths. This suggests that the models lack fundamental understanding of numbers and thus also maths. One approach to improve number representation is to scale up the vocabulary by having more individual number tokens. For example, GPT3 \cite{Brown-etal-2020-GPT3} has unique tokens from the numbers 0-520, whereas GPT4 \cite{OpenAI-GPT4-report-2023} has them up to 999. Despite general better performance of GPT4, it is not feasible to represent infinitely many numbers in finite memory capacity. Making the vocabulary larger also increases the computational costs.

\subsection{Tokenisation}
A more practical approach for representing all numbers is digit tokenisation \cite{spithourakis-riedel-2018-numeracy, geva-etal-2020-injecting} which separates numbers into a sequence of individual digits. This method improves upon conventional wordpiece tokenisation as shown with GenBERT \cite{geva-etal-2020-injecting} and Mistral-7B \cite{jiang-etal-2023-mistral} by reducing vocabulary size and ensuring precise representation of all numbers. Despite its advantages over conventional tokenisation algorithms, digit tokenisation has limitations. It relies on the model to aggregate digit embeddings into complete number representations.
During pretraining, models typically learn to aggregate subword tokens effectively for common words. However, not all numbers are encountered frequently enough during pretraining for the model to learn accurate aggregation. As an example, when the same question is posed with numbers represented differently (once as an integer and once scaled to the thousands), FLAN large with digit tokenisation shows a performance drop of 10\% \cite{sivakumar-moosavi-2023-fermat}. This indicates that the model struggles with numerical consistency and accurate aggregation of digit embeddings.

\subsection{Architectural level}
Change in model architecture also aids numerical reasoning as shown by NumNET \cite{ran-etal-2019-numnet} and xVAL \cite{golkar-etal-2023-xval}. NumNET extracts the numbers from the input question and passage to create a directed graph with relative magnitude information about each number present, e.g. which is greater than the others. After encoding the input question, this comparative information is also passed to the model to also be leveraged in answering the query.
Alternatively, xVAL generates two input encodings, a text only encoding where numbers are replaced by [NUM], and a number encoding with empty space for the text but the actual value of the number in each number's corresponding position. From the number preserving encoding, each number is converted to a vector with each entry as the number itself. The product of this vector with the embedding of [NUM] is then injected into the first layer of the transformer for each number in the input sequence. For decoding, a bespoke process is created to extract the predicted number instead of outputting the [NUM] token. 
Despite the positive contributions of these papers, their methods lack versatility as they are not adaptable off-the-shelf for any pretrained model.

\subsection{Loss Functions}
Another approach to improve numerical reasoning is for models to intrinsically learn better representation by introducing an inductive bias in the loss function. A simple approach is \citet{wallace-etal-2019-nlp}'s use of the mean squared error (MSE) loss across the batch to directly predict floats on a subset of DROP \cite{dua-etal-2019-drop} which consists of numerical answers. Unfortunately, this method is limited to datasets that only predict numbers.
Contrastive loss has also been used to manipulate the representation of numbers, for instance, \citet{petrak-etal-2023-arithmetic} draws nearer the representation generated by BPE and digit tokenisation of numbers through an auxiliary loss when doing extended pretraining to improve arithmetic reasoning in worded problems like DROP but also tables like SciGen \cite{moosavi-etal-2021-scigen}.  Similarly, \citet{li-etal-2022-seeking} use contrastive learning but on computation trees. They first generate computation trees for the mathematical operations. Then they use contrastive loss to pull nearer the graph representing the same operation, e.g. additions, and push other ones further. This is then integrated in the main loss and improves performance on two maths worded problem datasets, MathQA \cite{amini-etal-2019-mathqa} and Math23K \cite{wang-etal-2017-deep}. While these loss functions are adaptable with different models, contrastive training is computationally expensive and requires annotated data.

\subsection{Input Representation}
The most model agnostic method involves changing the representation of the numbers in the input text. \citet{wallace-etal-2019-nlp} explore worded forms of numbers, but this approach overly relies on the tokeniser which splits them into subwords. \citet{muffo-etal-2022-evaluating} decompose numbers into place values in reverse order, e.g. 123 = 3 units, 2 tens, 1 hundreds which helps when working with carry-on, e.g. when adding. However, this introduces many more tokens in the input which is undesirable, and also requires new vocabulary for each place value name. \citet{zhang-etal-2020-language-embeddings} convert all numbers into scientific notation, e.g. 314.1 is represented as 3141[EXP]2, improving models' ability to identify the magnitude of a number. Despite providing magnitudinal information, the number before [EXP] still needs to be represented faithfully. In fact, all the above strategies require the model to implicitly compute an overall aggregation for the numbers based on their individual components generated by the tokeniser of the model, whether these are digits or subwords.
A simple, yet effective method is to introduce pause tokens before predicting the answer \cite{goyal-etal-2024-pause}. This is evaluated by training a 1B parameter transformer model on C4 using [PAUSE] tokens and a 1\% improvement is shown on the numerical reasoning dataset, GSM8K \cite{cobbe-etal-2021-gsm8k}. While this method can be used for inference only, they conclude that pretraining is recommended, therefore less applicable to existing models.

Within this line of research, our work is more versatile. Unlike previous methods that rely on the model to implicitly learn aggregation, we focus on the explicit aggregation of digit embeddings using mathematical priors. This provides direct supervision for the aggregation process, improving the accuracy of number representation. Furthermore, our method ensures that the embedding for a given number aligns with its numerical neighbours, enhancing the model's numerical reasoning capabilities without altering the model architecture or requiring extensive retraining.

\section{Aggregation of Digit Embeddings}\label{Aggregation of digit embeddings}
Digit tokenisation has demonstrated its efficacy in enhancing numerical reasoning compared to BPE tokenisation. This improvement can be attributed to digit tokenisation's utilisation of pretrained embeddings for individual digits, allowing the model to learn the overall representation through contextualised embeddings. In contrast, BPE tends to fragment longer and less frequent numbers into random subsequences, resulting in less meaningful aggregations than those achieved through digit tokenisation. However, the implicit aggregation process employed by digit tokenisation remains unclear; specifically, how the model forms the overall aggregation of a number given the embeddings of its individual digits.

In this paper, we investigate a natural continuation of digit tokenisation, a mathematically motivated aggregation that takes into account the relative position of each digit within a number. Our approach generates an overall embedding for the number by considering the positional weight of each individual digit in that number. For example, given ``123'', the common understanding of numbers as base-10 is ``$1 \times 100 + 2 \times 10 + 3 \times 1$'', so the left most digits are weighted higher as they have a larger impact on the value of the number. 

We design our weighted scheme such that:
(1) the embeddings of single-digit numbers remain intact,  as these embeddings are effectively learned during pretraining, evidenced by the high performance of models on single-digit operations \citep{sivakumar-moosavi-2023-fermat}, (2) the weights of consecutive place values increase exponentially to reflect base-10, and (3) the weights do not sum to 1, meaning that it does not normalise the sum, allowing for numbers composed of the same digits, e.g. ``111'' and ``11'', to be represented differently. 
These properties would introduce a bias towards an accurate length of numbers and the correct digits from left to right as the left most digits are amplified, hence preserving natural numerical order.

We propose to calculate the weighted aggregated embedding $\boldsymbol{a}$ with  $a_i = \sum w_i \cdot d_i$ for $1 \le i \le N$ where $N$ is the number of digits, and the weights $w_i$ are defined as:
\begin{equation}
    w_i= 2^{N-i} \times \frac{3(N+1-i)(N+2-i)}{N(N+1)(N+2)}.
    \label{eqn:weight}
\end{equation}
\begin{figure}[!]
    \centering
    \includegraphics[width=0.49\textwidth]{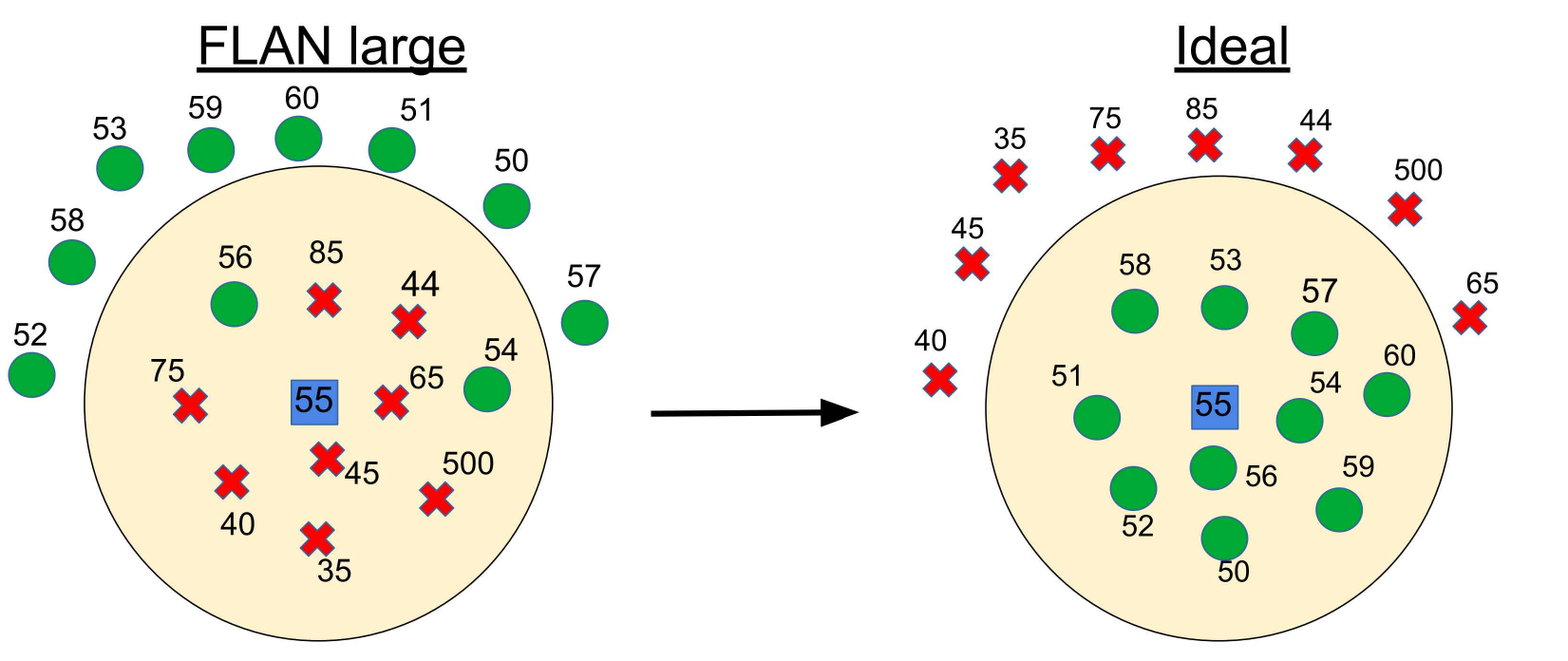}
    \caption{A 2D projection of the neighbourhood of the number token “55” in FLAN large is represented on the left. Ideally, number embeddings should reflect natural numerical proximity. In other words, the embedding for any given number should closely align with those of its immediate numerical neighbours, depicted on the right.}
    \label{fig:ekNN_vs_nkNN}
\end{figure}
These weights are designed to satisfy three key properties. \textbf{(1) Alignment with single-digit representations:} when $N=1$, $w_1=1$, ensuring compatibility with the model's pretraining on single digits. \textbf{(2) Exponential growth:} the exponential component $2^{N-i}$ mimics the base-10 system, providing an appropriate scale without causing the weights to grow too rapidly. This also ensures that the weights are not normalised. \textbf{(3) Regularisation Term:} the fractional component acts as a regularisation term, forming a normalised triangular number sequence. For instance, for a 3-digit number, the triangular sequence is 1,3,6, normalised to 0.1,0.3,0.6. This ensures that the difference between consecutive digit weights increases proportionally, i.e., $w_i-w_{i-1}=w_0 \times i$, replicating the exponential ratio between digit positions in a logarithmic space.
 
To validate the ability of an aggregated embedding to accurately represent numerical relationships, we use the F1-score to compare natural k-Nearest Neighbours ($n$kNN) with embedding k-Nearest Neighbours ($e$kNN). This comparison serves two purposes: firstly, to assess the embeddings' capacity to distinguish between distinct numbers, and secondly, to evaluate how well these embeddings mirror the natural numerical order. By defining $n$kNN as the set of mathematically adjacent numbers to a given integer $n$, and $e$kNN as the set of its closest neighbours in the embedding space, we create a direct measure of the embedding's effectiveness in preserving numerical proximity. The F1-score evaluates the alignment between $n$kNN and $e$kNN, penalising both the inclusion of incorrect neighbours and the omission of correct ones. A strong correlation between $n$kNN and $e$kNN, as reflected in a high F1-score, indicates that the embeddings faithfully capture the essence of numerical data as illustrated in Figure~\ref{fig:ekNN_vs_nkNN}.

\begin{figure}[t]
    \centering
    \includegraphics[width=0.49\textwidth]{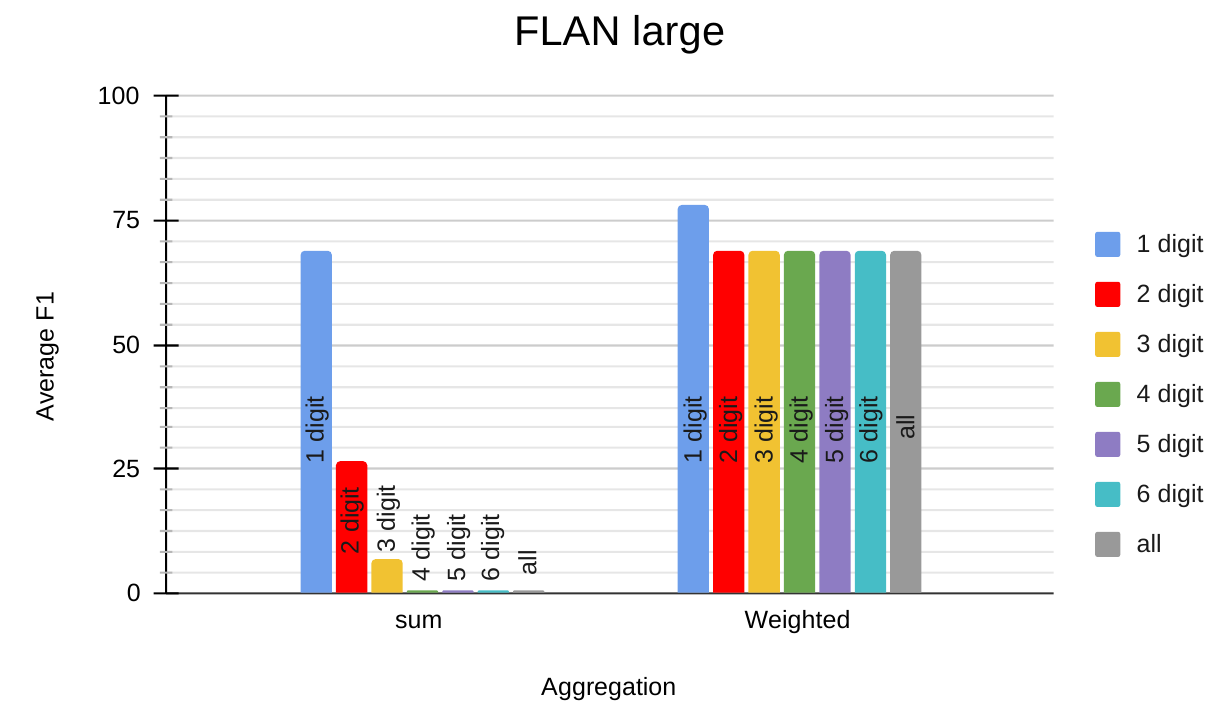}
    \caption{Average F1-score of FLAN large layer 1 numbers using sum and our weighted aggregation function with neighbourhood of 10.}
    \label{fig:NSNE_FLAN}
\end{figure}

We compare our bespoke weighted aggregation function to more standard aggregation functions such as sum. For a set of digit embeddings, we apply these functions along each dimension to generate a unique embedding for the number represented by these digits. Figure~\ref{fig:NSNE_FLAN} graphs the F1-score for our weighted function and sum over different digit lengths, i.e. 2-digit would be the numbers 10 to 99. Appendix~\ref{App: Agg functions} has the results for other aggregation functions: max, min, mean and median; these have the lowest alignment with natural order with an F1-score below 5\%. These functions all have a normalising property meaning that the length of the number has no bearing on the aggregated embedding, as the functions only retrieve one entry for each dimension therefore cases like ``1111'' would be equivalent to both ``11'' and ``1''. Contrastingly, sum has better F1-scores for up to 3 digits as it possesses magnitudinal information since all the entries are summed up for each dimension distinguishing, for instance, a 2-digit set from a 3-digit set as it simply adds more numbers. However, it is position agnostic - it assigns equal weight to all the digits irrespective of their relative positions. Therefore, the embeddings generated from permutations of the same digits will always be equivalent, e.g. ``85'' and ``58''. Since larger digit numbers have more such permutations, the F1-score reduces as the number of digits increases. Using this metric, the best aggregation is our weighted sum, the average F1-score rounds to 69\% for 2 digits onwards suggesting that our weighted sum is closer to the ideal depiction in Figure~\ref{fig:ekNN_vs_nkNN}. Undoubtedly, 1-digit F1-score is better as these embeddings are generated from pretraining, but also because the weighted scheme ensures that they are separated from the other number embeddings.

Despite this weighted scheme aligning the number embeddings with their natural order, the weights generated by Equation~\ref{eqn:weight} can surpass the precision used making it too large after a certain point. However, this behaviour is attenuated by the regularisation term which maintains the high F1-score of 69\% for, at least, up to 6-digit long numbers as shown in Figure~\ref{fig:NSNE_FLAN}. Theoretically, the newly formed number representation should contain  numbers that beginning with the same digits and only vary at the unit level. For example, the neighbourhood of 4523 should contain all numbers of the form 452X where X is a digit from 0 to 9, therefore eight of these coincide in the 10-nearest neighbour, namely 452\textbf{0}, 452\textbf{1}, 452\textbf{2}, 452\textbf{4}, 452\textbf{5}, 452\textbf{6}, 452\textbf{7}, and 452\textbf{8}.

\section{Integrating Aggregated Embeddings}
Given the construction of our mathematically grounded aggregation, we explore two distinct methodologies for enhancing numerical understanding in models, each targeting different aspects of number representation. The first method focuses on enriching the input data by integrating a mathematical aggregation directly into the input embedding as a special token. This approach requires no changes to the model's architecture, making it a flexible solution compatible with various models and suitable for a broad spectrum of tasks.

In contrast, the second approach aims to refine the model's output by improving how numbers are represented in the learned outcomes. This is achieved by incorporating the aggregation in the loss function, encouraging the model to generate number embeddings that align more closely to the correct numerical values. Specifically, this method includes an additional term in the loss calculation, which accounts for the distance between the aggregated embedding of the predicted number and that of the true number. This targeted intervention is particularly effective in tasks requiring precise numerical predictions, helping the model develop a more nuanced and accurate representation of numbers.

The baseline implementation for both methods is the same as \citet{petrak-etal-2023-arithmetic} with digit tokenisation surrounded by [F] and [/F] tokens to mark the start and end of the number identified using the regular expression ``(\textbackslash d*\textbackslash.)?\textbackslash d+''.
\subsection{Aggregation in Input Embeddings}
In our first approach, we enhance the input embedding by incorporating the computed aggregation directly. This is achieved by first digitising numbers and delineating them with special tokens as done by \citet{petrak-etal-2023-arithmetic}. Additionally, we introduce a special token, [AGG], positioned as follows where $d_i$ represent the digit tokens: [F] [AGG] [$d_1$] ... [$d_n$] [/F]. The embedding for this [AGG] token is initialised with the aggregation of the digit embeddings based on Equation~\ref{eqn:weight}.

\subsection{Aggregation in Loss Function}
Language generation models typically use a cross-entropy loss function ($\mathcal{L}_{CE}$) \citep{lewis-etal-2020-bart, raffel-etal-2020-T5}.
To improve the model's ability to predict numbers accurately, we introduce an auxiliary loss ($\mathcal{L}_{AUX}$) to calculate the mean squared error between the aggregate embedding of the gold and predicted numbers. Understanding and predicting numbers is inherently more complex than predicting a single word or sub-word because they consist of multiple digits, each carrying different significance. For example, in answering the question ``Mary's salary is £900 a month, but she pays £579 in rent. How much salary does she have left?'', the answers 320, 230, 32, or 456 are all incorrect. However, 320 is more accurate compared to others because its magnitude is closer to the correct answer, 321. Incorporating this new auxiliary loss would help the model predict digits that are closer to the gold answer, enhancing its precision in numerical predictions by recognising the relative significance of each digit within a number.

Given a prediction $p$ and the gold label $l$, we compute the weighted sum of the digits\footnote{Should the answers not be numerical, the model is penalise by arbitrarily setting $\mathcal{L}_{AUX}$ to $20$.} for both $p$ and $l$. This process generates two single embedding representations: $W(p)$ for the prediction, and $W(l)$ for the gold label. The distance between these two embeddings is then calculated using the log\footnote{Log base 2 is used to regularise the auxiliary loss.} mean squared error (equivalent to the euclidean distance):
\begin{equation}
    \mathcal{L}_{AUX} = \log_2 \left( \, \lVert W(p)-W(l) \lVert_2 \, \right)
    \label{eqn:loss_aux}
\end{equation}
Finally, the two losses are linearly interpolated by a hyperparameter, $\lambda$:
\begin{equation}
    \mathcal{L} =  \lambda \times \mathcal{L}_{CE}  + (1-\lambda) \times  \mathcal{L}_{AUX}
    \label{eqn:loss}
\end{equation}

\begin{table*}[t]
\resizebox{\textwidth}{!}{
\centering
\begin{tabular}{|cl|c|cccccccccccccc|}
\hline
\multicolumn{2}{|c|}{} &
   &
  \multicolumn{14}{c|}{FERMAT} \\ \cline{4-17} 
\multicolumn{2}{|c|}{\multirow{-6.5}{*}{\begin{tabular}[c]{@{}c@{}}Incorporating Weights\\ (Accuracy \%)\end{tabular}}} &
 \multicolumn{1}{c|}{\multirow{-3.5}{*}{\begin{turn}{90} MAWPS\end{turn}}} &
  \multicolumn{1}{c}{\begin{turn}{90}Original\end{turn}} &
  \multicolumn{1}{c}{\begin{turn}{90}Commuted\end{turn}} &
  \multicolumn{1}{c}{\begin{turn}{90}Integers 0 to 1000 \;\end{turn}} &
  \multicolumn{1}{c}{\begin{turn}{90}2-digit integers\end{turn}} &
  \multicolumn{1}{c}{\begin{turn}{90}3-digit integers\end{turn}} &
  \multicolumn{1}{c}{\begin{turn}{90}4-digit integers\end{turn}} &
  \multicolumn{1}{c}{\begin{turn}{90}1000+\end{turn}} &
  \multicolumn{1}{c}{\begin{turn}{90}1000+ same\end{turn}} &
  \multicolumn{1}{c}{\begin{turn}{90}1dp random\end{turn}} &
  \multicolumn{1}{c|}{\begin{turn}{90}2dp random\end{turn}} &
  \multicolumn{1}{c}{\begin{turn}{90}a+b\end{turn}} &
  \multicolumn{1}{c}{\begin{turn}{90}a-b\end{turn}} &
  \multicolumn{1}{c}{\begin{turn}{90}a*b\end{turn}} &
  \multicolumn{1}{c|}{\begin{turn}{90}a/b\end{turn}}\\ \hline
\multicolumn{1}{|c|}{} &
  Digits &
  \cellcolor[HTML]{B2B2B2}19.20 &
  \cellcolor[HTML]{B2B2B2}16.65 &
  \cellcolor[HTML]{B2B2B2}8.73 &
  \cellcolor[HTML]{B2B2B2}10.26 &
  \cellcolor[HTML]{B2B2B2}13.41 &
  \cellcolor[HTML]{B2B2B2}10.89 &
  \cellcolor[HTML]{B2B2B2}7.74 &
  \cellcolor[HTML]{B2B2B2}5.58 &
  \cellcolor[HTML]{B2B2B2}10.89 &
  \cellcolor[HTML]{B2B2B2}17.82 &
  \multicolumn{1}{c|}{\cellcolor[HTML]{B2B2B2}8.37} &
  \cellcolor[HTML]{B2B2B2}40.91 &
  \cellcolor[HTML]{B2B2B2}10.62 &
  \cellcolor[HTML]{B2B2B2}9.56 &
  \cellcolor[HTML]{B2B2B2}11.76 \\
\multicolumn{1}{|c|}{} &
  {[}AGG{]} + Digits &
  \cellcolor[HTML]{AFD095}\textbf{+2.00} &
  \cellcolor[HTML]{DDE8CB}+0.63 &
  \cellcolor[HTML]{AFD095}+1.53 &
  \cellcolor[HTML]{FF6D6D}-1.17 &
  \cellcolor[HTML]{FFD7D7}-0.90 &
  \cellcolor[HTML]{FF6D6D}-2.16 &
  \cellcolor[HTML]{FFD7D7}-0.27 &
  \cellcolor[HTML]{DDE8CB}+0.09 &
  \cellcolor[HTML]{DDE8CB}\textbf{+0.09} &
  \cellcolor[HTML]{AFD095}\textbf{+1.08} &
  \multicolumn{1}{c|}{\cellcolor[HTML]{FFD7D7}-0.27} &
  \cellcolor[HTML]{FF6D6D}-3.90 &
  \cellcolor[HTML]{FFD7D7}-0.74 &
  \cellcolor[HTML]{AFD095}+1.77 &
  0.00 \\
\multicolumn{1}{|c|}{\multirow{-3}{*}{\begin{tabular}[c]{@{}c@{}}BART base \\ (140M)\end{tabular}}} &
  Digits + Aux Loss &
  \cellcolor[HTML]{AFD095}+1.40 &
  \cellcolor[HTML]{AFD095}\textbf{+1.89} &
  \cellcolor[HTML]{AFD095}\textbf{+1.80} &
  \cellcolor[HTML]{DDE8CB}\textbf{+0.54} &
  \cellcolor[HTML]{DDE8CB}\textbf{+0.81} &
  0.00 &
  \cellcolor[HTML]{DDE8CB}\textbf{+0.81} &
  \cellcolor[HTML]{AFD095}\textbf{+1.17} &
  \cellcolor[HTML]{FF6D6D}-1.26 &
  \cellcolor[HTML]{DDE8CB}+0.18 &
  \multicolumn{1}{c|}{\cellcolor[HTML]{DDE8CB}\textbf{+0.63}} &
  \cellcolor[HTML]{AFD095}\textbf{+2.01} &
  \cellcolor[HTML]{DDE8CB}\textbf{+0.19} &
  \cellcolor[HTML]{AFD095}\textbf{+4.25} &
  \cellcolor[HTML]{FF6D6D}-1.27 \\ \hline
\multicolumn{1}{|c|}{} &
  Digits &
  \cellcolor[HTML]{B2B2B2}23.00 &
  \cellcolor[HTML]{B2B2B2}28.35 &
  \cellcolor[HTML]{B2B2B2}17.82 &
  \cellcolor[HTML]{B2B2B2}17.10 &
  \cellcolor[HTML]{B2B2B2}22.86 &
  \cellcolor[HTML]{B2B2B2}17.37 &
  \cellcolor[HTML]{B2B2B2}13.77 &
  \cellcolor[HTML]{B2B2B2}10.35 &
  \cellcolor[HTML]{B2B2B2}18.72 &
  \cellcolor[HTML]{B2B2B2}25.83 &
  \multicolumn{1}{c|}{\cellcolor[HTML]{B2B2B2}18.45} &
  \cellcolor[HTML]{B2B2B2}63.38 &
  \cellcolor[HTML]{B2B2B2}19.57 &
  \cellcolor[HTML]{B2B2B2}12.92 &
  \cellcolor[HTML]{B2B2B2}11.27 \\
\multicolumn{1}{|c|}{} &
  {[}AGG{]} + Digits &
  \cellcolor[HTML]{DDE8CB}+0.80 &
  \cellcolor[HTML]{AFD095}\textbf{+2.79} &
  \cellcolor[HTML]{DDE8CB}+0.27 &
  \cellcolor[HTML]{AFD095}+2.52 &
  \cellcolor[HTML]{DDE8CB}+0.81 &
  \cellcolor[HTML]{AFD095}\textbf{+1.80} &
  \cellcolor[HTML]{AFD095}\textbf{+2.79} &
  \cellcolor[HTML]{AFD095}\textbf{+1.80} &
  \cellcolor[HTML]{DDE8CB}+0.90 &
  \cellcolor[HTML]{DDE8CB}+0.45 &
  \multicolumn{1}{c|}{\cellcolor[HTML]{FFD7D7}-0.09} &
  \cellcolor[HTML]{AFD095}\textbf{+4.48} &
  \cellcolor[HTML]{AFD095}+3.21 &
  \cellcolor[HTML]{FFD7D7}-0.27 &
  \cellcolor[HTML]{AFD095}+1.08 \\
\multicolumn{1}{|c|}{\multirow{-3}{*}{\begin{tabular}[c]{@{}c@{}}FLAN base \\ (250M)\end{tabular}}} &
  Digits + Aux Loss &
  \cellcolor[HTML]{AFD095}\textbf{+1.80} &
  \cellcolor[HTML]{AFD095}+2.25 &
  \cellcolor[HTML]{DDE8CB}\textbf{+0.36} &
  \cellcolor[HTML]{AFD095}\textbf{+3.15} &
  \cellcolor[HTML]{AFD095}\textbf{+2.16} &
  \cellcolor[HTML]{AFD095}+1.71 &
  \cellcolor[HTML]{AFD095}\textbf{+2.79} &
  \cellcolor[HTML]{DDE8CB}+0.81 &
  \cellcolor[HTML]{AFD095}\textbf{+3.87} &
  \cellcolor[HTML]{AFD095}\textbf{+1.89} &
  \multicolumn{1}{c|}{\cellcolor[HTML]{FFD7D7}-0.18} &
  \cellcolor[HTML]{AFD095}+3.90 &
  \cellcolor[HTML]{AFD095}\textbf{+5.80} &
  \cellcolor[HTML]{DDE8CB}\textbf{+0.27} &
  \cellcolor[HTML]{AFD095}\textbf{+1.57} \\ \hline
\multicolumn{1}{|c|}{} &
  Digits &
  \cellcolor[HTML]{B2B2B2}28.80 &
  \cellcolor[HTML]{B2B2B2}42.39 &
  \cellcolor[HTML]{B2B2B2}21.06 &
  \cellcolor[HTML]{B2B2B2}25.65 &
  \cellcolor[HTML]{B2B2B2}31.32 &
  \cellcolor[HTML]{B2B2B2}24.30 &
  \cellcolor[HTML]{B2B2B2}21.87 &
  \cellcolor[HTML]{B2B2B2}16.47 &
  \cellcolor[HTML]{B2B2B2}23.31 &
  \cellcolor[HTML]{B2B2B2}36.36 &
  \multicolumn{1}{c|}{\cellcolor[HTML]{B2B2B2}25.83} &
  \cellcolor[HTML]{B2B2B2}63.12 &
  \cellcolor[HTML]{B2B2B2}39.88 &
  \cellcolor[HTML]{B2B2B2}18.23 &
  \cellcolor[HTML]{B2B2B2}18.14 \\
\multicolumn{1}{|c|}{} &
  {[}AGG{]} + Digits &
  \cellcolor[HTML]{AFD095}\textbf{+1.20} &
  \cellcolor[HTML]{DDE8CB}+0.45 &
  \cellcolor[HTML]{DDE8CB}\textbf{+0.45} &
  \cellcolor[HTML]{DDE8CB}+0.81 &
  \cellcolor[HTML]{AFD095}+2.07 &
  \cellcolor[HTML]{AFD095}\textbf{+2.79} &
  \cellcolor[HTML]{DDE8CB}\textbf{+0.99} &
  \cellcolor[HTML]{AFD095}+1.35 &
  \cellcolor[HTML]{AFD095}\textbf{+2.88} &
  \cellcolor[HTML]{DDE8CB}+0.27 &
  \multicolumn{1}{c|}{\cellcolor[HTML]{DDE8CB}+0.54} &
  \cellcolor[HTML]{AFD095}\textbf{+6.17} &
  \cellcolor[HTML]{AFD095}\textbf{+3.83} &
  \cellcolor[HTML]{DDE8CB}\textbf{+0.53} &
  \cellcolor[HTML]{AFD095}\textbf{+1.47} \\
\multicolumn{1}{|c|}{\multirow{-3}{*}{\begin{tabular}[c]{@{}c@{}}FLAN large\\ (780M)\end{tabular}}} &
  Digits + Aux Loss &
  \cellcolor[HTML]{DDE8CB}+1.00 &
  \cellcolor[HTML]{DDE8CB}\textbf{+0.99} &
  \cellcolor[HTML]{FFD7D7}-0.18 &
  \cellcolor[HTML]{AFD095}\textbf{+1.62} &
  \cellcolor[HTML]{AFD095}\textbf{+2.88} &
  \cellcolor[HTML]{AFD095}\textbf{+2.79} &
  \cellcolor[HTML]{DDE8CB}+0.72 &
  \cellcolor[HTML]{AFD095}\textbf{+1.53} &
  \cellcolor[HTML]{AFD095}+1.26 &
  \cellcolor[HTML]{AFD095}\textbf{+1.26} &
  \multicolumn{1}{c|}{\cellcolor[HTML]{DDE8CB}\textbf{+0.63}} &
  \cellcolor[HTML]{FFD7D7}-0.39 &
  \cellcolor[HTML]{AFD095}+1.79 &
  \cellcolor[HTML]{DDE8CB}+0.18 &
  \cellcolor[HTML]{FF6D6D}-1.08 \\ \hline
\end{tabular}
}
\caption{Results change in Accuracy from baseline after including aggregate embeddings in input embedding ([AGG] + Digits) and auxiliary loss (Digits + Aux Loss) for BART base, FLAN base and FLAN large. Darker shades of green and red indicate an absolute change greater than 1\%.}
\label{tab: main}
\end{table*}

\begin{table*}[h!]
\resizebox{\textwidth}{!}{%
\begin{tabular}{|cl|c|cccccccccccccc|}
\hline
\multicolumn{2}{|c|}{} &
   &
  \multicolumn{14}{c|}{FERMAT} \\ \cline{4-17} 
\multicolumn{2}{|c|}{\multirow{-6.5}{*}{\begin{tabular}[c]{@{}c@{}}Incorporating Weights\\ (CER \%)\end{tabular}}} &
    \multicolumn{1}{c|}{\multirow{-3.5}{*}{\begin{turn}{90} MAWPS\end{turn}}} &
  \multicolumn{1}{c}{\begin{turn}{90}Original\end{turn}} &
  \multicolumn{1}{c}{\begin{turn}{90}Commuted\end{turn}} &
  \multicolumn{1}{c}{\begin{turn}{90}Integers 0 to 1000 \;\end{turn}} &
  \multicolumn{1}{c}{\begin{turn}{90}2-digit integers\end{turn}} &
  \multicolumn{1}{c}{\begin{turn}{90}3-digit integers\end{turn}} &
  \multicolumn{1}{c}{\begin{turn}{90}4-digit integers\end{turn}} &
  \multicolumn{1}{c}{\begin{turn}{90}1000+\end{turn}} &
  \multicolumn{1}{c}{\begin{turn}{90}1000+ same\end{turn}} &
  \multicolumn{1}{c}{\begin{turn}{90}1dp random\end{turn}} &
  \multicolumn{1}{c|}{\begin{turn}{90}2dp random\end{turn}} &
  \multicolumn{1}{c}{\begin{turn}{90}a+b\end{turn}} &
  \multicolumn{1}{c}{\begin{turn}{90}a-b\end{turn}} &
  \multicolumn{1}{c}{\begin{turn}{90}a*b\end{turn}} &
  \multicolumn{1}{c|}{\begin{turn}{90}a/b\end{turn}} \\ \hline
\multicolumn{1}{|c|}{} &
  Digits &
  \cellcolor[HTML]{B2B2B2}77.73 &
  \cellcolor[HTML]{B2B2B2}89.59 &
  \cellcolor[HTML]{B2B2B2}90.32 &
  \cellcolor[HTML]{B2B2B2}72.87 &
  \cellcolor[HTML]{B2B2B2}71.93 &
  \cellcolor[HTML]{B2B2B2}72.25 &
  \cellcolor[HTML]{B2B2B2}74.04 &
  \cellcolor[HTML]{B2B2B2}77.01 &
  \cellcolor[HTML]{B2B2B2}50.29 &
  \cellcolor[HTML]{B2B2B2}54.42 &
  \multicolumn{1}{c|}{\cellcolor[HTML]{B2B2B2}62.23} &
  \cellcolor[HTML]{B2B2B2}50.31 &
  \cellcolor[HTML]{B2B2B2}74.12 &
  \cellcolor[HTML]{B2B2B2}60.73 &
  \cellcolor[HTML]{B2B2B2}75.51 \\
\multicolumn{1}{|c|}{} &
  [AGG] + Digits &
  \cellcolor[HTML]{AFD095}\textbf{-1.79} &
  \cellcolor[HTML]{AFD095}\textbf{-12.40} &
  \cellcolor[HTML]{DDE8CB}\textbf{-0.83} &
  \cellcolor[HTML]{FFD7D7}+0.46 &
  \cellcolor[HTML]{FFD7D7}+0.51 &
  \cellcolor[HTML]{FF6D6D}+1.19 &
  \cellcolor[HTML]{DDE8CB}-0.16 &
  \cellcolor[HTML]{DDE8CB}-0.44 &
  \cellcolor[HTML]{FFD7D7}+0.94 &
  \cellcolor[HTML]{AFD095}-1.38 &
  \multicolumn{1}{c|}{\cellcolor[HTML]{AFD095}-1.28} &
  \cellcolor[HTML]{FF6D6D}+3.08 &
  \cellcolor[HTML]{AFD095}\textbf{-1.58} &
  \cellcolor[HTML]{FF6D6D}+1.21 &
  \cellcolor[HTML]{AFD095}\textbf{-2.22} \\
\multicolumn{1}{|c|}{\multirow{-3}{*}{\begin{tabular}[c]{@{}c@{}}BART base \\ (140M)\end{tabular}}} &
  Digits + Aux Loss &
  \cellcolor[HTML]{FFD7D7}+0.76 &
  \cellcolor[HTML]{AFD095}-1.88 &
  \cellcolor[HTML]{DDE8CB}-0.53 &
  \cellcolor[HTML]{FFD7D7}+0.17 &
  \cellcolor[HTML]{FFD7D7}+0.20 &
  \cellcolor[HTML]{FFD7D7}+0.34 &
  \cellcolor[HTML]{AFD095}\textbf{-1.06} &
  \cellcolor[HTML]{DDE8CB}\textbf{-0.53} &
  \cellcolor[HTML]{AFD095}\textbf{-1.89} &
  \cellcolor[HTML]{AFD095}\textbf{-1.59} &
  \multicolumn{1}{c|}{\cellcolor[HTML]{AFD095}\textbf{-1.78}} &
  \cellcolor[HTML]{AFD095}\textbf{-2.45} &
  \cellcolor[HTML]{DDE8CB}-0.23 &
  \cellcolor[HTML]{AFD095}\textbf{-2.75} &
  \cellcolor[HTML]{FFD7D7}+0.26 \\ \hline
\multicolumn{1}{|c|}{} &
  Digits &
  \cellcolor[HTML]{B2B2B2}67.71 &
  \cellcolor[HTML]{B2B2B2}75.32 &
  \cellcolor[HTML]{B2B2B2}169.52 &
  \cellcolor[HTML]{B2B2B2}67.37 &
  \cellcolor[HTML]{B2B2B2}67.68 &
  \cellcolor[HTML]{B2B2B2}67.94 &
  \cellcolor[HTML]{B2B2B2}67.86 &
  \cellcolor[HTML]{B2B2B2}68.86 &
  \cellcolor[HTML]{B2B2B2}50.95 &
  \cellcolor[HTML]{B2B2B2}43.77 &
  \multicolumn{1}{c|}{\cellcolor[HTML]{B2B2B2}47.80} &
  \cellcolor[HTML]{B2B2B2}39.84 &
  \cellcolor[HTML]{B2B2B2}87.81 &
  \cellcolor[HTML]{B2B2B2}60.96 &
  \cellcolor[HTML]{B2B2B2}91.52 \\
\multicolumn{1}{|c|}{} &
  [AGG] + Digits &
  \cellcolor[HTML]{DDE8CB}-0.98 &
  \cellcolor[HTML]{AFD095}\textbf{-1.40} &
  \cellcolor[HTML]{DDE8CB}-0.29 &
  \cellcolor[HTML]{AFD095}\textbf{-1.11} &
  \cellcolor[HTML]{AFD095}\textbf{-1.41} &
  \cellcolor[HTML]{AFD095}\textbf{-1.19} &
  \cellcolor[HTML]{AFD095}\textbf{-1.67} &
  \cellcolor[HTML]{DDE8CB}-0.96 &
  \cellcolor[HTML]{FF6D6D}+1.26 &
  \cellcolor[HTML]{AFD095}-1.33 &
  \multicolumn{1}{c|}{\cellcolor[HTML]{DDE8CB}\textbf{-0.39}} &
  \cellcolor[HTML]{AFD095}\textbf{-1.64} &
  \cellcolor[HTML]{AFD095}-1.94 &
  \cellcolor[HTML]{DDE8CB}-0.17 &
  \cellcolor[HTML]{DDE8CB}-0.50 \\
\multicolumn{1}{|c|}{\multirow{-3}{*}{\begin{tabular}[c]{@{}c@{}}FLAN base \\ (250M)\end{tabular}}} &
  Digits + Aux Loss &
  \cellcolor[HTML]{AFD095}\textbf{-1.54} &
  \cellcolor[HTML]{DDE8CB}-0.83 &
  \cellcolor[HTML]{AFD095}\textbf{-1.09} &
  \cellcolor[HTML]{AFD095}-1.09 &
  \cellcolor[HTML]{AFD095}-1.15 &
  \cellcolor[HTML]{DDE8CB}-0.80 &
  \cellcolor[HTML]{AFD095}-1.39 &
  \cellcolor[HTML]{AFD095}\textbf{-1.23} &
  \cellcolor[HTML]{AFD095}\textbf{-2.09} &
  \cellcolor[HTML]{AFD095}\textbf{-1.82} &
  \multicolumn{1}{c|}{\cellcolor[HTML]{DDE8CB}-0.30} &
  \cellcolor[HTML]{AFD095}-1.25 &
  \cellcolor[HTML]{AFD095}\textbf{-3.15} &
  \cellcolor[HTML]{DDE8CB}\textbf{-0.72} &
  \cellcolor[HTML]{DDE8CB}\textbf{-0.93} \\ \hline
\multicolumn{1}{|c|}{} &
  Digits &
  \cellcolor[HTML]{B2B2B2}63.13 &
  \cellcolor[HTML]{B2B2B2}69.71 &
  \cellcolor[HTML]{B2B2B2}76.46 &
  \cellcolor[HTML]{B2B2B2}63.02 &
  \cellcolor[HTML]{B2B2B2}62.69 &
  \cellcolor[HTML]{B2B2B2}63.53 &
  \cellcolor[HTML]{B2B2B2}63.96 &
  \cellcolor[HTML]{B2B2B2}66.67 &
  \cellcolor[HTML]{B2B2B2}49.90 &
  \cellcolor[HTML]{B2B2B2}37.63 &
  \multicolumn{1}{c|}{\cellcolor[HTML]{B2B2B2}42.31} &
  \cellcolor[HTML]{B2B2B2}39.00 &
  \cellcolor[HTML]{B2B2B2}58.84 &
  \cellcolor[HTML]{B2B2B2}52.84 &
  \cellcolor[HTML]{B2B2B2}70.49 \\
\multicolumn{1}{|c|}{} &
  [AGG] + Digits &
  \cellcolor[HTML]{AFD095}-2.57 &
  \cellcolor[HTML]{AFD095}\textbf{-44.77} &
  \cellcolor[HTML]{AFD095}\textbf{-10.81} &
  \cellcolor[HTML]{AFD095}-1.02 &
  \cellcolor[HTML]{DDE8CB}-0.10 &
  \cellcolor[HTML]{AFD095}\textbf{-1.63} &
  \cellcolor[HTML]{DDE8CB}-0.65 &
  \cellcolor[HTML]{DDE8CB}-0.89 &
  \cellcolor[HTML]{FF6D6D}+1.78 &
  \cellcolor[HTML]{DDE8CB}-0.93 &
  \multicolumn{1}{c|}{\cellcolor[HTML]{AFD095}-1.23} &
  \cellcolor[HTML]{AFD095}\textbf{-6.16} &
  \cellcolor[HTML]{AFD095}\textbf{-7.80} &
  \cellcolor[HTML]{AFD095}\textbf{-5.49} &
  \cellcolor[HTML]{AFD095}\textbf{-7.19} \\
\multicolumn{1}{|c|}{\multirow{-3}{*}{\begin{tabular}[c]{@{}c@{}}FLAN large\\ (780M)\end{tabular}}} &
  Digits + Aux Loss &
  \cellcolor[HTML]{AFD095}\textbf{-3.45} &
  \cellcolor[HTML]{AFD095}-45.42 &
  \cellcolor[HTML]{AFD095}-2.72 &
  \cellcolor[HTML]{AFD095}\textbf{-1.20} &
  \cellcolor[HTML]{DDE8CB}\textbf{-0.24} &
  \cellcolor[HTML]{AFD095}-1.09 &
  \cellcolor[HTML]{AFD095}\textbf{-1.23} &
  \cellcolor[HTML]{AFD095}\textbf{-1.31} &
  \cellcolor[HTML]{AFD095}\textbf{-2.57} &
  \cellcolor[HTML]{AFD095}\textbf{-1.11} &
  \multicolumn{1}{c|}{\cellcolor[HTML]{AFD095}\textbf{-1.27}} &
  \cellcolor[HTML]{AFD095}-3.47 &
  \cellcolor[HTML]{AFD095}-6.14 &
  \cellcolor[HTML]{AFD095}-2.93 &
  \cellcolor[HTML]{AFD095}-4.74 \\ \hline
\end{tabular}}
\caption{Results in Character Error Rate (CER) as a percentage over the target string with change from baseline after including aggregate embeddings in input embedding ([AGG] + Digits) and auxiliary loss (Digits + Aux Loss) for BART base, FLAN base and FLAN large. With CER, lower CER indicates a better performance, green highlight reduced CER i.e. negative change, and red the opposite. Darker shades of green and red indicate an absolute change greater than 1\%.}
\label{tab: CER}
\end{table*}

\section{Experimental Setup}
Both methods are evaluated on two different pretrained models, BART base (140M) \cite{lewis-etal-2020-bart} and FLAN base (250M) \cite{wei-etal-2022-FLAN}. Additionally, we evaluate on FLAN large (780M) to explore the effect of model size. BART is an encoder-decoder pre-trained on five corrupted document tasks from books and Wikipedia data. FLAN is an instruction-finetuned version of T5 \cite{raffel-etal-2020-T5} which is trained on C4 using transfer learning.

We evaluate our proposed methods on two different test sets: FERMAT \cite{sivakumar-moosavi-2023-fermat}, and MAWPS \cite{koncel-kedziorski-etal-2016-mawps}. Both FERMAT and MAWPS consist of English maths worded problem that can be tackled by BART and FLAN as shown by \citet{sivakumar-moosavi-2023-fermat} and where the answer is a single number. This enables us to evaluate our method strictly on numerical outputs reducing the interference of other difficulties such as predicting words and units, or extracting spans. FERMAT is a multi-view evaluation set which has different test sets with different number representations while keeping the maths problem fixed. The different test sets distinguish different number types of which we select the ones that separate integers into digit length (2-digit, 3-digit, 4-digit), contain a mixture of integers less than 1000, contain a mixture of integers greater than 1000, the sets of one and two decimal place numbers, and a test set that takes the original set and scales the number to more than 4-digit numbers; these allow us to evaluate which number representation the models support better. FERMAT's training set is augmented from different templates making it independent to its test sets. MAWPS, on the other hand, has the same domain for both training and testing. It is a widely used dataset to evaluate numerical reasoning, chiefly because it is small and easy to train with small models. We finetune the models on each dataset's respective training data (see Appendix~\ref{App: Datasets}) using the hyperparameters described in Appendix~\ref{App: Hyperparameters}.

Accuracy is the general metric used to evaluate these datasets, however, since it is sometimes too stringent and neglects to reflect some improvements of the model, we also use a variation of edit distance \cite{Levenshtein-1966-edit-distance} as a supplementary metric. Edit distance helps see improvement in the predictions despite being incorrect; it calculates how many insertions, deletions or substitutions is required for the prediction to be transformed into the gold label number on a string level. In this paper, we will use Character Error Rate (CER) which is a character level (digit level) edit distance  as a percentage over the string length of the target. The lower the CER, the closer the prediction is to the gold label.

\section{Impact of Integrating Aggregations}
Table~\ref{tab: main} presents the results of our exploration into the effects of integrating mathematical aggregation into the three models across two distinct settings. The bold values indicate the stronger improvement between the two incorporation strategies. For the majority of the test splits, the strongest performance of the examined models is observed when the aggregation is incorporated into the auxiliary loss. This suggests that incorporating aggregation at the output level is more effective than incorporating it in the input embedding. However, this may be due to the fact that adding a new token in the input might require more than just fine-tuning, such as an extended pretraining phase. This aligns with the observations made by \citet{goyal-etal-2024-pause}, who found that the addition of the pause token only became effective from pretraining.

FLAN large, on the other hand, has a more balanced performance but an overall higher improvement when the aggregation is incorporated in the input as shown particularly from all the green cells in the row [AGG] + Digits. Therefore, a certain model size may be required to learn a new token and leverage the information it provides. This reinforces that an aggregated embedding provides useful signal to improve number understanding but how it is integrated is also crucial.

For the operations, the improvements is generally positive across all of them, however, evidently greater for addition and subtraction than multiplication and division. This resonates with the fact that digits positions are more informative for the first two operations, especially when, for instance, aligning them to perform calculations.

When focusing on smaller integers (columns ``Integers 0 to 1000'' to ``4-digit integers''), incorporating the weighted embedding in the auxiliary loss consistently yields better performance, with all cells being green and showing the highest scores. For smaller integers, models likely already possess a strong implicit representation, making the explicit [AGG] token less impactful. However, at the decoding stage, the auxiliary loss enhances precision by penalising incorrect predictions.

\begin{table*}[t!]
\resizebox{\textwidth}{!}{
\begin{tabular}{|cl|c|cccccccccccccc|}
\hline
\multicolumn{2}{|c|}{} &
  \multicolumn{1}{l|}{} &
  \multicolumn{14}{c|}{FERMAT} \\ \cline{4-17} 
\multicolumn{2}{|c|}{\multirow{-6.5}{*}{\begin{tabular}[c]{@{}c@{}}Aggregated Embedding\\ (Accuracy \%)\end{tabular}}} &
  \multicolumn{1}{c|}{\multirow{-3.5}{*}{\begin{turn}{90} MAWPS\end{turn}}} &
  \multicolumn{1}{c}{\begin{turn}{90}Original\end{turn}} &
  \multicolumn{1}{c}{\begin{turn}{90}Commuted\end{turn}} &
  \multicolumn{1}{c}{\begin{turn}{90}Integers 0 to 1000 \;\end{turn}} &
  \multicolumn{1}{c}{\begin{turn}{90}2-digit integers\end{turn}} &
  \multicolumn{1}{c}{\begin{turn}{90}3-digit integers\end{turn}} &
  \multicolumn{1}{c}{\begin{turn}{90}4-digit integers\end{turn}} &
  \multicolumn{1}{c}{\begin{turn}{90}1000+\end{turn}} &
  \multicolumn{1}{c}{\begin{turn}{90}1000+ same\end{turn}} &
  \multicolumn{1}{c}{\begin{turn}{90}1dp random\end{turn}} &
  \multicolumn{1}{c|}{\begin{turn}{90}2dp random\end{turn}} &
  \multicolumn{1}{c}{\begin{turn}{90}a+b\end{turn}} &
  \multicolumn{1}{c}{\begin{turn}{90}a-b\end{turn}} &
  \multicolumn{1}{c}{\begin{turn}{90}a*b\end{turn}} &
  \multicolumn{1}{c|}{\begin{turn}{90}a/b\end{turn}} \\ \hline
\multicolumn{1}{|c|}{} &
  Digits &
  \cellcolor[HTML]{B2B2B2}19.20 &
  \cellcolor[HTML]{B2B2B2}16.65 &
  \cellcolor[HTML]{B2B2B2}8.73 &
  \cellcolor[HTML]{B2B2B2}10.26 &
  \cellcolor[HTML]{B2B2B2}13.41 &
  \cellcolor[HTML]{B2B2B2}10.89 &
  \cellcolor[HTML]{B2B2B2}7.74 &
  \cellcolor[HTML]{B2B2B2}5.58 &
  \cellcolor[HTML]{B2B2B2}10.89 &
  \cellcolor[HTML]{B2B2B2}17.82 &
  \multicolumn{1}{c|}{\cellcolor[HTML]{B2B2B2}8.37} &
  \cellcolor[HTML]{B2B2B2}40.91 &
  \cellcolor[HTML]{B2B2B2}10.62 &
  \cellcolor[HTML]{B2B2B2}9.56 &
  \cellcolor[HTML]{B2B2B2}11.76 \\
\multicolumn{1}{|c|}{} &
  Digits + [AGG] &
  \cellcolor[HTML]{FF6D6D}-1.40 &
  \cellcolor[HTML]{FF6D6D}-14.76 &
  \cellcolor[HTML]{FF6D6D}-7.74 &
  \cellcolor[HTML]{FF6D6D}-8.82 &
  \cellcolor[HTML]{FF6D6D}-10.98 &
  \cellcolor[HTML]{FF6D6D}-8.73 &
  \cellcolor[HTML]{FF6D6D}-6.75 &
  \cellcolor[HTML]{FF6D6D}-5.58 &
  \cellcolor[HTML]{FF6D6D}-10.35 &
  \cellcolor[HTML]{FF6D6D}-14.76 &
  \multicolumn{1}{c|}{\cellcolor[HTML]{FF6D6D}-7.83} &
  \cellcolor[HTML]{FF6D6D}-36.82 &
  \cellcolor[HTML]{FF6D6D}-9.38 &
  \cellcolor[HTML]{FF6D6D}-8.94 &
  \cellcolor[HTML]{FF6D6D}-9.51 \\
\multicolumn{1}{|c|}{} &
  [AGG] + Digits &
  \cellcolor[HTML]{AFD095}\textbf{+2.00} &
  \cellcolor[HTML]{DDE8CB}\textbf{+0.63} &
  \cellcolor[HTML]{AFD095}\textbf{+1.53} &
  \cellcolor[HTML]{FF6D6D}-1.17 &
  \cellcolor[HTML]{FFD7D7}-0.90 &
  \cellcolor[HTML]{FF6D6D}-2.16 &
  \cellcolor[HTML]{FFD7D7}0.27 &
  \cellcolor[HTML]{DDE8CB}\textbf{+0.09} &
  \cellcolor[HTML]{DDE8CB}\textbf{+0.09} &
  \cellcolor[HTML]{AFD095}\textbf{+1.08} &
  \multicolumn{1}{c|}{\cellcolor[HTML]{FFD7D7}0.27} &
  \cellcolor[HTML]{FF6D6D}3.90 &
  \cellcolor[HTML]{FFD7D7}0.74 &
  \cellcolor[HTML]{AFD095}\textbf{+1.77} &
  0.00 \\
\multicolumn{1}{|c|}{\multirow{-4}{*}{\begin{tabular}[c]{@{}c@{}}BART base \\ (140M)\end{tabular}}} &
   [PAUSE] + Digits &
  \cellcolor[HTML]{FF6D6D}-1.40 &
  \cellcolor[HTML]{DDE8CB}+0.18 &
  \cellcolor[HTML]{FFD7D7}-0.45 &
  \cellcolor[HTML]{FFD7D7}-0.18 &
  \cellcolor[HTML]{FFD7D7}-0.63 &
  \cellcolor[HTML]{FFD7D7}-0.90 &
  \cellcolor[HTML]{FFD7D7}-0.36 &
  \cellcolor[HTML]{FFD7D7}-0.27 &
  \cellcolor[HTML]{FF6D6D}-3.87 &
  \cellcolor[HTML]{FFD7D7}-0.90 &
  \multicolumn{1}{c|}{0.00} &
  \cellcolor[HTML]{FF6D6D}-8.51 &
  \cellcolor[HTML]{FFD7D7}-0.31 &
  \cellcolor[HTML]{AFD095}+1.68 &
  \cellcolor[HTML]{FF6D6D}-2.06 \\
  \hline
\multicolumn{1}{|c|}{} &
  Digits &
  \cellcolor[HTML]{B2B2B2}23.00 &
  \cellcolor[HTML]{B2B2B2}28.35 &
  \cellcolor[HTML]{B2B2B2}17.82 &
  \cellcolor[HTML]{B2B2B2}17.10 &
  \cellcolor[HTML]{B2B2B2}22.86 &
  \cellcolor[HTML]{B2B2B2}17.37 &
  \cellcolor[HTML]{B2B2B2}13.77 &
  \cellcolor[HTML]{B2B2B2}10.35 &
  \cellcolor[HTML]{B2B2B2}18.72 &
  \cellcolor[HTML]{B2B2B2}25.83 &
  \multicolumn{1}{c|}{\cellcolor[HTML]{B2B2B2}18.45} &
  \cellcolor[HTML]{B2B2B2}63.38 &
  \cellcolor[HTML]{B2B2B2}19.57 &
  \cellcolor[HTML]{B2B2B2}12.92 &
  \cellcolor[HTML]{B2B2B2}11.27 \\
\multicolumn{1}{|c|}{} &
    Digits + [AGG] &
  \cellcolor[HTML]{AFD095}\textbf{+1.80} &
  \cellcolor[HTML]{FF6D6D}-1.53 &
  \cellcolor[HTML]{FF6D6D}-2.07 &
  \cellcolor[HTML]{DDE8CB}+0.99 &
  \cellcolor[HTML]{FF6D6D}-1.89 &
  \cellcolor[HTML]{FFD7D7}-0.36 &
  \cellcolor[HTML]{DDE8CB}+0.63 &
  \cellcolor[HTML]{AFD095}+1.35 &
  \cellcolor[HTML]{FFD7D7}-0.63 &
  \cellcolor[HTML]{FF6D6D}-1.98 &
  \multicolumn{1}{c|}{\cellcolor[HTML]{FFD7D7}-0.99} &
  \cellcolor[HTML]{DDE8CB}+0.45 &
  \cellcolor[HTML]{AFD095}+3.89 &
  \cellcolor[HTML]{FF6D6D}-2.39 &
  \cellcolor[HTML]{FFD7D7}-0.10 \\
\multicolumn{1}{|c|}{} &
  [AGG] + Digits &
  \cellcolor[HTML]{DDE8CB}+0.80 &
  \cellcolor[HTML]{AFD095}\textbf{+2.79} &
  \cellcolor[HTML]{DDE8CB}\textbf{+0.27} &
  \cellcolor[HTML]{AFD095}\textbf{+2.52} &
  \cellcolor[HTML]{DDE8CB}+0.81 &
  \cellcolor[HTML]{AFD095}\textbf{+1.80} &
  \cellcolor[HTML]{AFD095}\textbf{+2.79} &
  \cellcolor[HTML]{AFD095}+1.80 &
  \cellcolor[HTML]{DDE8CB}+0.90 &
  \cellcolor[HTML]{DDE8CB}+0.45 &
  \multicolumn{1}{c|}{\cellcolor[HTML]{FFD7D7}-0.09} &
  \cellcolor[HTML]{AFD095}\textbf{+4.48} &
  \cellcolor[HTML]{AFD095}+3.21 &
  \cellcolor[HTML]{FFD7D7}-0.27 &
  \cellcolor[HTML]{AFD095}+1.08 \\
\multicolumn{1}{|c|}{\multirow{-4}{*}{\begin{tabular}[c]{@{}c@{}}FLAN base \\ (250M)\end{tabular}}} &
  [PAUSE] + Digits &
  \cellcolor[HTML]{DDE8CB}+1.00 &
  \cellcolor[HTML]{AFD095}+2.07 &
  \cellcolor[HTML]{FFD7D7}-0.54 &
  \cellcolor[HTML]{AFD095}+1.98 &
  \cellcolor[HTML]{AFD095}\textbf{+1.44} &
  \cellcolor[HTML]{AFD095}\textbf{+1.80} &
  \cellcolor[HTML]{AFD095}+2.61 &
  \cellcolor[HTML]{AFD095}\textbf{+2.52} &
  \cellcolor[HTML]{AFD095}\textbf{+2.16} &
  \cellcolor[HTML]{AFD095}\textbf{+2.61} &
  \multicolumn{1}{c|}{\cellcolor[HTML]{AFD095}\textbf{+1.71}} &
  \cellcolor[HTML]{AFD095}+3.18 &
  \cellcolor[HTML]{AFD095}\textbf{+5.99} &
  \cellcolor[HTML]{FF6D6D}\-1.95 &
  \cellcolor[HTML]{AFD095}\textbf{+3.43} \\
 \hline
\multicolumn{1}{|c|}{} &
  Digits &
  \cellcolor[HTML]{B2B2B2}28.80 &
  \cellcolor[HTML]{B2B2B2}42.39 &
  \cellcolor[HTML]{B2B2B2}21.06 &
  \cellcolor[HTML]{B2B2B2}25.65 &
  \cellcolor[HTML]{B2B2B2}31.32 &
  \cellcolor[HTML]{B2B2B2}24.30 &
  \cellcolor[HTML]{B2B2B2}21.87 &
  \cellcolor[HTML]{B2B2B2}16.47 &
  \cellcolor[HTML]{B2B2B2}23.31 &
  \cellcolor[HTML]{B2B2B2}36.36 &
  \multicolumn{1}{c|}{\cellcolor[HTML]{B2B2B2}25.83} &
  \cellcolor[HTML]{B2B2B2}63.12 &
  \cellcolor[HTML]{B2B2B2}39.88 &
  \cellcolor[HTML]{B2B2B2}18.23 &
  \cellcolor[HTML]{B2B2B2}18.14 \\
\multicolumn{1}{|c|}{} &
    Digits + [AGG] &
  \cellcolor[HTML]{FF6D6D}-2.80 &
  \cellcolor[HTML]{FF6D6D}-2.16 &
  \cellcolor[HTML]{AFD095}\textbf{+1.35} &
  \cellcolor[HTML]{AFD095}\textbf{+1.89} &
  \cellcolor[HTML]{AFD095}+1.08 &
  \cellcolor[HTML]{AFD095}+1.44 &
  \cellcolor[HTML]{AFD095}+1.62 &
  \cellcolor[HTML]{AFD095}+2.16 &
  \cellcolor[HTML]{AFD095}\textbf{+5.40} &
  \cellcolor[HTML]{FF6D6D}-1.17 &
  \multicolumn{1}{c|}{\cellcolor[HTML]{DDE8CB}+0.54} &
  \cellcolor[HTML]{AFD095}\textbf{+8.57} &
  \cellcolor[HTML]{FF6D6D}-8.15 &
  \cellcolor[HTML]{FFD7D7}-0.97 &
  \cellcolor[HTML]{AFD095}+1.18 \\ 
\multicolumn{1}{|c|}{} &
  [AGG] + Digits &
  \cellcolor[HTML]{AFD095}\textbf{+1.20} &
  \cellcolor[HTML]{DDE8CB}\textbf{+0.45} &
  \cellcolor[HTML]{DDE8CB}+0.45 &
  \cellcolor[HTML]{DDE8CB}+0.81 &
  \cellcolor[HTML]{AFD095}+2.07 &
  \cellcolor[HTML]{AFD095}+2.79 &
  \cellcolor[HTML]{DDE8CB}+0.99 &
  \cellcolor[HTML]{AFD095}+1.35 &
  \cellcolor[HTML]{AFD095}+2.88 &
  \cellcolor[HTML]{DDE8CB}+0.27 &
  \multicolumn{1}{c|}{\cellcolor[HTML]{DDE8CB}+0.54} &
  \cellcolor[HTML]{AFD095}+6.17 &
  \cellcolor[HTML]{AFD095}\textbf{+3.83} &
  \cellcolor[HTML]{DDE8CB}\textbf{+0.53} &
  \cellcolor[HTML]{AFD095}+1.47 \\
\multicolumn{1}{|c|}{\multirow{-4}{*}{\begin{tabular}[c]{@{}c@{}}FLAN large\\ (780M)\end{tabular}}} &
[PAUSE] + Digits &
  \cellcolor[HTML]{FF6D6D}-1.40 &
  \cellcolor[HTML]{FFD7D7}-0.45 &
  \cellcolor[HTML]{FFD7D7}-0.45 &
  \cellcolor[HTML]{AFD095}\textbf{+1.89} &
  \cellcolor[HTML]{AFD095}\textbf{+3.69} &
  \cellcolor[HTML]{AFD095}\textbf{+2.88} &
  \cellcolor[HTML]{AFD095}\textbf{+3.06} &
  \cellcolor[HTML]{AFD095}\textbf{+2.25} &
  \cellcolor[HTML]{AFD095}+5.04 &
  \cellcolor[HTML]{AFD095}\textbf{+1.17} &
  \multicolumn{1}{c|}{\cellcolor[HTML]{AFD095}\textbf{+2.61}} &
  \cellcolor[HTML]{AFD095}+6.17 &
  \cellcolor[HTML]{AFD095}+1.17 &
  \cellcolor[HTML]{FF6D6D}-1.77 &
  \cellcolor[HTML]{AFD095}\textbf{+3.53} \\
\hline
\end{tabular}}
\caption{Comparing the aggregated embedding at the input level with a pause token and positioning the token after the digits. Darker shades of green and red indicate an absolute change greater than 1\%.}
\label{tab: agg tokens}
\end{table*}

For the 1000+ columns, using accuracy, the pattern is not evident, however, Table~\ref{tab: CER} presents the character error rate (CER) comparing both incorporating strategies for all three models, and highlights that using the auxiliary loss clearly reduces the CER more than explicitly using the aggregation in the input. The auxiliary loss encourages the model to predict the correct answer as the CER is lower. However, since the weights assigned to each digit position is lower as it gets closer to the units, the auxiliary accounts less for it, reducing precision. As a consequence, despite the CER reducing, since the entire number is not predicted correctly, improvement fails to be reflected in the accuracy.


\section{Analysis of Aggregation Embedding in the Input}

The first integration method relies on prepending the aggregated embedding token, [AGG], before the digits. The position of the token is before what it represents, similar in nature to BERT's \cite{devlin-etal-2019-bert} [CLS] token, which is an aggregation token of the entire input. However, \citet{goyal-etal-2024-pause} use a [PAUSE] token posteriori to the digit tokens to act as processing time after concluding that prepending it had less impact. Consequently, we also evaluate our proposed method by appending the aggregation token, i.e. Digits + [AGG]. Table~\ref{tab: agg tokens} clearly shows that this configuration for both base models underperforms compared to [AGG] + Digit as rows have more red entries. In fact, it performs worse than the baseline with only digit tokenisation. For FLAN large, the results between [AGG] prepended and appended are closer to one another, but prepended, the impact is positive for each test set and on average better by 1\% than [AGG] used posteriori. Seeing the token before the digits might provide magnitude information of the overall number which would indicate the importance of each digit to come, whereas having it after might interfere with the representation that the model has already started to create implicitly from seeing the digits first. 

Additionally, we test the impact of providing the aggregated token by replacing it with a randomly initialised [PAUSE] token akin to \citet{goyal-etal-2024-pause}. From Table~\ref{tab: agg tokens}, we observe that for BART, neither [AGG], nor [PAUSE] have a great positive impact on the performance. This confirms that BART struggles to learn new tokens from finetuning alone. The FLAN models are more adaptable to the new tokens as seen by the greener rows. However, the overwhelming bold entries with the [PAUSE] token indicate that both FLAN base and large perform better with a [PAUSE] token acting as a blank space for the model to process the information. It is possible that that the model uses this token to create an implicit representation of the number. Nevertheless, the average improvement between the [PAUSE] and [AGG] differs by less than 0.5\% implying that a different aggregation function or a full hyperparameter search could reverse the trend.

\section{Future Work}
Our proposed aggregation strategy has shown encouraging steps towards better number representation. However, as with observations made in previous work, the effect of new strategies report minimal improvement on smaller models but greater impact on larger models \cite{cobbe-etal-2021-gsm8k, wei-etal-2022-chain-of-thought}. Therefore, an evaluation of our proposed method on larger scale models would verify the scalability of this approach.

The weighting scheme, presented in Equation~\ref{eqn:weight}, offers a straightforward method for aggregating digit embeddings. However, as numbers increase in length, their aggregated embeddings tend to drift away from the original numerical embedding subspace. This divergence could be addressed by enabling the model to adapt to this new embedding space by exploring extended pretraining, or constructing weighting schemes that remain closer to the numerical subspace while satisfying the criteria outlined in Section~\ref{Aggregation of digit embeddings}.

Our auxiliary loss, grounded in Mean Squared Error, shows promising results for penalising the model's erroneous predictions and nudging it towards more accurate outcomes. Given that the values resulting from standard cross-entropy and the MSE of the aggregated embeddings may span vastly different value ranges, crafting a loss function that aligns more closely in magnitude with the output of cross-entropy could mitigate the risk of exerting excessive regularisation pressure.

\section{Conclusion}
Improving numerical reasoning is a challenging task, increasing model sizes or focusing on data augmentation helps but at the cost of a substantial additional training time or computations. Digit tokenisation has been a pioneering work in improving how models encode and decode numbers, however the aggregation of the digit is done implicitly. We advance this idea by explicitly providing an aggregated number embedding that is more mathematically sound. These embeddings are generated as weighted sums of the digit embeddings by accounting for the digits relative position in the number. We then incorporate them in two model agnostic forms: in the input level as an additional token, and in an auxiliary MSE loss. Our promising results demonstrate that, as a proof-of-concept, even a straightforward aggregation with simple incorporation techniques can positively impact number understanding. Therefore, testing it at a larger scale, developing sophisticated aggregation functions, and refining the integration of the auxiliary loss presents valuable avenues for future research.

\section{Limitations}
Some of the limitations of this work are discussed in the Future Work section. However, we give details of further limitations relating to the size of the models used, and the compatibility and growth of our proposed weighted aggregation function.
\\ Due to financial and resource constraints the hypothesis that the methods for incorporating the aggregated embedding in larger architectures would lead to greater performance based on the improvement observed on smaller model is not verified.
\\ In addition, while the weighted scheme is designed using mathematical priors, it is specifically created for integers, therefore it may not be compatible with decimals or alternative representation of numbers such as 01 for 1. Nonetheless, from Table~\ref{tab: CER}, we note that CER reduces for both 1dp and 2dp therefore our aggregated embedding method has promising scope for all numbers.
\\ Furthermore, the weight function described in Equation~\ref{eqn:weight} does not converge, therefore for a sufficiently large number of digit it would grow beyond the accuracy provided by the model. However, we explain in Section~\ref{Aggregation of digit embeddings} with the aid of Figure~\ref{fig:NSNE_FLAN} that, for up to 6-digits, the weighted scheme functions well with no signs of deterioration. Moreover, in natural text, very large numbers tend to be shorten using a more appropriate unit, for example, the world population of 8114693010 is more often expressed as 8 billion reducing the numbers of digits needed considerably. But this raises the question of predicting the correct unit which would lead to future work.
\\ Nonetheless, our weighting scheme leverages digit embedding, therefore it is heavily dependent on them, particularly on the relative distance of the digit embedding to one another. In FLAN large, the embedding of the digit 0 is more distant from the embeddings of other digits, which causes it to frequently include numbers ending in 0 when the target contains a 0, or to exclude them otherwise. As explained in Section~\ref{Aggregation of digit embeddings}, the optimal neighbour would include all numbers with different unit value, and this alone would achieve an F1-score of at least 70\%.
\\ Lastly, the experiments were conducted using a single random seed. While this ensures consistency and reproducibility, having access to better resources would have enabled us to run the experiments with multiple seeds. This would have allowed us to calculate the average improvement achieved by using aggregated digit embeddings to represent numbers.

\section*{Acknowledgements}
This work was supported by the Centre for Doctoral Training in Speech and Language Technologies (SLT) and their Applications funded by UK Research and Innovation [grant number EP/S023062/1]. We also acknowledge IT Services at The University of Sheffield for the provision of services for High Performance Computing. Additional thanks to the reviewers for their encouraging comments and discussion, and particularly to Danae Sanchez Villegas, Mugdha Pandya, Valeria Pastorino, Huiyin Xue and Constantinos Karouzos for their continued feedback in through the research.

\bibliography{anthology,custom}
\bibliographystyle{acl_natbib}

\appendix

\section*{Appendix}
\label{sec:appendix}
\section{Aggregation functions}\label{App: Agg functions}
\begin{figure*}
    \centering
    \includegraphics[width=\textwidth]{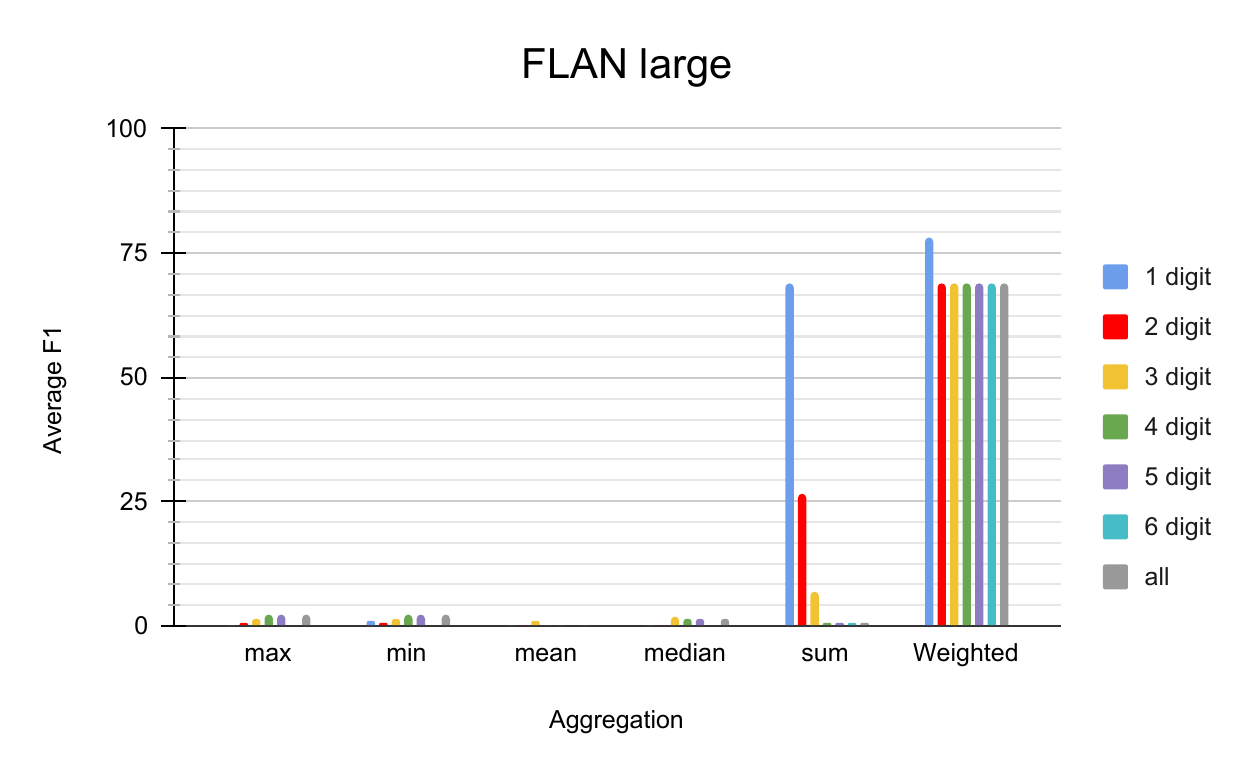}
    \caption{Average F1-score of FLAN large layer 1 numbers using max, min, median, mean sum and our weighted aggregation function with neighbourhood of 10.}
    \label{fig:NSNE_FLAN_All}
\end{figure*}
Figure~\ref{fig:NSNE_FLAN_All} shows that F1-score for numbers with up to 6-digits across six different aggregation functions. The F1-score for max, min, mean and median are all below 5\%.

\section{Datasets}\label{App: Datasets}
The datasets' split is given in Table~\ref{tab: dataset_info}. MAWPS is a dataset generated by combining different ones ranging from addition and subtraction to simultaneous equations. The collation of questions is split to create the train, development and test set. FERMAT is a large dataset which has a training and development set automatically generated from 100 templates using different numbers from the following four categories: small integers (less than 1000), large integers (between 1000 and 100000), 1 decimal place and 2 decimal place numbers. The test set is independently generated from two maths worded problem datasets, and then augmented to create 21 test sets of which we use 11.
\begin{table}[h!]
\centering
    \begin{tabular}{|l|ccc|}
        \hline
        \multicolumn{1}{|c|}{Datasets} & \multicolumn{1}{c}{Train} & \multicolumn{1}{c}{Dev} & \multicolumn{1}{c|}{Test} \\ \hline
        MAWPS & 1500 & 373 & \multicolumn{1}{c|}{500} \\
        FERMAT & 200000 & 1000 & 1111x11 \\
        \hline
    \end{tabular}
    
    \caption{Train, development, and test splits of MAWPS and FERMAT.}
    \label{tab: dataset_info}
\end{table}

\section{Hyperparameters}\label{App: Hyperparameters}
All experiments were conducted using an Nvidia Tesla A100 with 80G and with a weight decay of 0.005, warm-up of 100, float32 and 3 generation beams, max input length = 128, max target length=16, and seed=42. Due to limited computational resources, a full grid search of hyperparameter was impossible, however, we do a lambda search in the range 0.4 to 0.8 in 0.05 increments. Specific hyperparameters as well as computation time for dataset and model combinations can be found in Table~\ref{tab: Hyperparameters}.

\begin{table*}[t]
\centering
\begin{tabular}{|c|c|c|c|c|c|c|}
\hline
Datasets &
  Models &
  \multicolumn{1}{c|}{\begin{tabular}[c]{@{}c@{}}Learning  Rate\end{tabular}} &
  Epochs &
  \begin{tabular}[c]{@{}c@{}}Batch Size\end{tabular} &
  Lambda &
  \begin{tabular}[c]{@{}c@{}}Training Time\end{tabular} \\ \hline
\multirow{3}{*}{MAWPS}  & \begin{tabular}[c]{@{}c@{}}BART base\end{tabular}  & \multirow{3}{*}{1.00E-04} & 150 & 128 & 0.6  & 1h   \\ \cline{2-2} \cline{4-7} 
                        & \begin{tabular}[c]{@{}c@{}}FLAN base\end{tabular}  &                           & 150 & 64  & 0.6  & 1h   \\ \cline{2-2} \cline{4-7} 
                        & \begin{tabular}[c]{@{}c@{}}FLAN large\end{tabular} &                           & 100 & 16  & 0.65 & 1.5h \\ \hline
\multirow{3}{*}{FERMAT} & \begin{tabular}[c]{@{}c@{}}BART base\end{tabular}  & \multirow{3}{*}{1.00E-05} & 50  & 128 & 0.6  & 37h  \\ \cline{2-2} \cline{4-7} 
                        & \begin{tabular}[c]{@{}c@{}}FLAN base\end{tabular}  &                           & 50  & 64  & 0.65 & 48h  \\ \cline{2-2} \cline{4-7} 
                        & \begin{tabular}[c]{@{}c@{}}FLAN large\end{tabular} &                           & 50  & 16  & 0.4  & 87h  \\ \hline
\end{tabular}
\caption{Specific hyperparameters for MAWPS and FERMAT based on the models trained. Training time is also provided as a rounded figure.}
\label{tab: Hyperparameters}
\end{table*}

\end{document}